\documentclass[letterpaper, 10 pt, conference]{ieeeconf}  
\IEEEoverridecommandlockouts        % This command is only needed if you want to use the \thanks command
\usepackage{microtype}
\usepackage{times} % assumes new font selection scheme installed
\usepackage{amsmath} % assumes amsmath package installed
\usepackage{amssymb}  % assumes amsmath package installed
\usepackage{multirow}
\usepackage[compress,sort]{cite}

\usepackage[hidelinks,pdftex,pdfauthor={Justin Miller, Jonathan P.\ How},pdftitle={Demand Estimation and Chance-Constrained Fleet Management for Ride Hailing}]{hyperref}

\usepackage[authormarkuptext=name,addedmarkup=bf,authormarkupposition=left]{changes}
\definechangesauthor[name={J.~H.}, color={blue}]{jh}
\definechangesauthor[name={J.~M.}, color={orange}]{jm}
\definechangesauthor[name={S.~C.}, color={purple}]{sc}
\setremarkmarkup{\bf(#2)}

\title{\LARGE \bf Demand Estimation and Chance-Constrained Fleet Management for Ride Hailing}

\author{Justin Miller and Jonathan P.\ How% <-this % stops a space
  \thanks{Laboratory of Information and Decision Systems,
    Massachusetts Institute of Technology, 77 Massachusetts Ave.,
    Cambridge, MA, USA {\tt\small \{justinm, jhow\}@mit.edu}}%
}

\usepackage{mathtools}

\usepackage[labelfont=bf,font=small]{caption}
\usepackage{float}
\usepackage[labelfont=bf,font=small]{subcaption}

\usepackage{amssymb}

\usepackage[linesnumbered,ruled,vlined]{algorithm2e}

\DeclareMathOperator*{\argmin}{arg\,\!min}
\DeclareMathOperator*{\argmax}{arg\,\!max}

\usepackage{balance}

\usepackage[capitalize]{cleveref}
\crefformat{equation}{(#1)}
\Crefformat{equation}{(#1)}
\crefname{equation}{}{}
\Crefname{equation}{}{}
\Crefname{table}{Table}{Tables}
\Crefname{figure}{Figure}{Figures}

\allowdisplaybreaks
\makeatletter
\renewcommand\paragraph{\@startsection{subsubsection}{4}{\z@}%
	{0.5ex \@plus1ex \@minus.2ex}%
	{-.15em}%
	{\normalfont\normalsize\itshape}}
\makeatother

\def\Tau{{\rm T}}

\begin{document}

\maketitle
\thispagestyle{empty}
\pagestyle{empty}

%%%%%%%%%%%%%%%%%%%%%%%%%%%%%%%%%%%%%%%%%%%%%%%%%%%%%%%%%%%%%%%%%%%%%%%%%%%%%%%%
\begin{abstract}
In autonomous Mobility on Demand (MOD) systems, customers request rides from a fleet of shared vehicles that can be automatically positioned in response to customer demand.
Recent approaches to MOD systems have focused on environments where customers can only request rides through an app or by waiting at a station.
This paper develops MOD fleet management approaches for ride hailing, where customers may instead request rides simply by hailing a passing vehicle, an approach of particular importance for campus MOD systems.
The challenge for ride hailing is that customer demand is not explicitly provided as it would be with an app, but rather customers are only served if a vehicle happens to be located at the arrival location.
This work focuses on maximizing the number of served hailing customers in an MOD system by learning and utilizing customer demand.
A Bayesian framework is used to define a novel customer demand model which incorporates observed pedestrian traffic to estimate customer arrival locations with a quantification of uncertainty.
An exploration planner is proposed which routes MOD vehicles in order to reduce arrival rate uncertainty.
A robust ride hailing fleet management planner is proposed which routes vehicles under the presence of uncertainty using a chance-constrained formulation.
Simulation of a real-world MOD system on MIT's campus demonstrates the effectiveness of the planners.
The customer demand model and exploration planner are demonstrated to reduce estimation error over time and the ride hailing planner is shown to improve the fraction of served customers in the system by 73\% over a baseline exploration approach.
\end{abstract}

%%%%%%%%%%%%%%%%%%%%%%%%%%%%%%%%%%%%%%%%%%%%%%%%%%%%%%%%%%%%%%%%%%%%%%%%%%%%%%%%
\section{Introduction}\label{sec:introduction}
\begin{figure}
	\center
	\includegraphics[width=1\columnwidth] {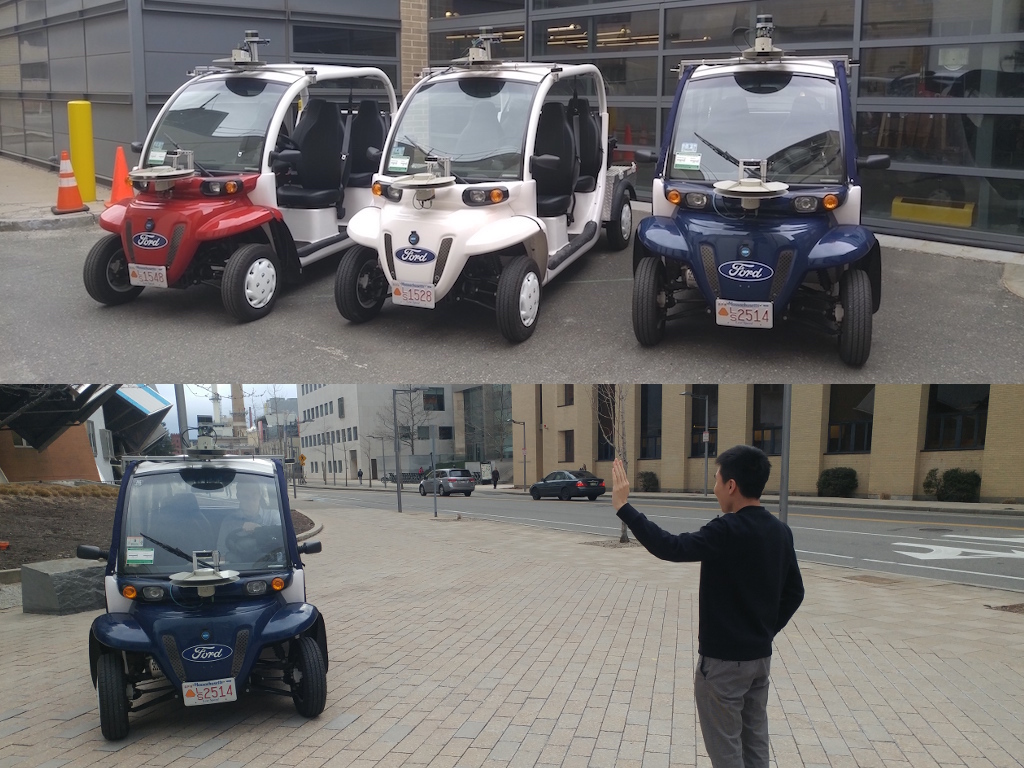}
	\caption{A fleet of three electric shuttles operates on MIT campus. The shuttles are manually driven but equipped with camera and Lidar for sensing pedestrians. Customers may hail rides directly from any nearby shuttle.}
	\label{fig::fleet}
\end{figure}

%WHAT is the problem?
%WHY is it interesting and important?
A paradigm shift in transportation is coming by way of the autonomous vehicle, which promises to save lives and reduce costs with fewer accidents \cite{anderson_autonomous_2014}.
Autonomous vehicles are expected to first be deployed within a Mobility On Demand (MOD) setting, as evidenced by companies such as Uber \cite{ross_uber_2016}, nuTonomy \cite{ackerman_hail_2017}, and Ford \cite{_ford_????} currently in active development of autonomous MOD fleets.
Modern approaches to autonomous MOD systems focus on two ways for customers to request rides: either customers walk to a station where they expect to find a parked car, or they request a ride using an app.
However, traditional taxi MOD systems have allowed a third option where customers can $\textit{hail}$ a ride by simply waving down a passing vehicle; an option not accounted for in existing autonomous MOD systems.
This work is motivated by the MIT MOD system which operates in a campus environment and allows customers to both make app requests and hail rides directly from a fleet of electric shuttles shown in \Cref{fig::fleet}.
Experience has shown that while customers using the MOD system have reduced commute times on campus, the hassle of requesting a ride via an app often causes many customers to default to walking.
Ride hailing can lower the barrier to entry by allowing customers who plan to walk, to instead hail a ride if they encounter an MOD vehicle.
The challenge for ride hailing is that the MOD system is not provided a customer arrival location as it would be with an app.
Unless a vehicle is in proximity of a customer's arrival location, that customer will not be served.
Furthermore, there are often more potential customer arrival locations than MOD vehicles available to monitor them.
For an MOD system to be successful, the distribution of MOD vehicles across potential arrival locations must be managed to match the customer demand \cite{mitchell_mobility_2008}.

%WHY is it hard? (E.g., why do naive approaches fail?)
The two main challenges for effective MOD management are estimating customer demand and managing vehicles with respect to the estimated demand.
Typical customer demand modeling approaches focus on the use of historical data consisting of locations and times that customers were served in the past.
If customers are submitting requests via an app, then historical customer arrival data can easily be obtained from request logs.
As more requests arrive via the app, the customer demand model can be improved.
But for customers only requesting rides via hailing, then historical customer arrivals are only obtained if vehicles were at the right place at the right time to encounter newly arrived customers.
A trial-and-error approach to assigning vehicles to park and wait sequentially at every potential arrival location would be slow and result in missed customers.
Furthermore, estimation of future customer arrival locations in real-time will inherently result in uncertainty for the predicted demand.
If uncertainty is not accounted for, vehicles may be assigned to wait at locations where no customers are arriving, again resulting in missed customers elsewhere within the MOD system.

%WHAT are the key components of my approach and results? Also include any specific limitations.
This work addresses the challenges of estimating customer demand and managing an autonomous vehicle fleet in order to improve the number of served customers in a campus MOD system.
A key component of the approach is the use of pedestrian traffic to estimate customer demand without the need for historical databases or uninformed trial-and-error assignment.
In a campus environment, areas with a large amount of pedestrian traffic are likely to be locations with potential customers.
The approach assumes customer arrivals to be correlated with total pedestrian arrivals due to campus-related events such as lunch breaks and regular schedules.
In \cite{miller_dynamic_2016}, a method for measuring pedestrian arrival rates using camera and Lidar sensors onboard MOD shuttles is introduced. 
This work extends that method by developing a customer demand model based on correlations with measured pedestrian arrival rates.
An exploration planner is developed which routes vehicles throughout the MOD network in order to minimize the uncertainty in pedestrian arrival rates.
The method is demonstrated to improve the estimates of customer arrival rates as vehicles explore the network.
Additionally, the work focuses on quantifying the uncertainty in arrival rate estimates using a Bayesian framework.
The arrival rate probabilities are incorporated into robust ride hailing policies which plan under uncertainty in customer arrival. 
Expected value and chance-constrained MOD fleet planners are demonstrated to improve the number of served customers. 
The concepts and approaches for the paper are demonstrated in the accompanying video available at \url{https://youtu.be/PW9snGPoohs}.

%WHAT ARE THE CONTRIBUTIONS
The contributions of this work are: 
1) modeling customer demand through the use of real-time pedestrian measurements made from MOD vehicle sensors; 
2) quantifying and reducing uncertainty in ride hailing customer arrivals through MOD fleet management; 
and 
3) increasing the number of served hailing customers through MOD fleet planning under uncertainty.

\section{Related Work}\label{sec:related_work}
%WHY hasn't it been solved before? (Or, what's wrong with previous proposed solutions? How does mine differ?)
Several automated MOD planners have been proposed, each focused on specific operating environments.
MOD operation can be classified into car sharing, ride request, ride sharing, and ride hailing environments.
In car sharing environments, customers walk to an origin hub and drive vehicles to a destination hub. 
As vehicles service customers, vehicles will be moved across the network and may be far from the next customer arrival location, causing an imbalance between future customer demand and fleet locations.
Car sharing planners focus on using known customer arrival rates to rebalance  vehicles across the network \cite{pavone_robotic_2012,volkov_markov-based_2012,zhang_control_2016,zhang_model_2016}.
These methods assume accurate knowledge of customer arrival rates, and do not address measurement uncertainty.
In \cite{miao_data-driven_2016}, customer arrival uncertainty is addressed through a robust optimization, with focus on rebalancing fleets on city-wide scales.

In ride request environments, customers request rides using an app and a vehicle is routed to them.
Ride sharing environments are an extension of ride request environments where vehicle capacities are increased and different customers may be serviced simultaneously by sharing a ride.
Ride request and ride hailing planners focus on optimizing the assignment of customers to vehicles in order to improve customer quality of service \cite{fagnant_dynamic_2015,miller_predictive_2016}.
These methods assume that customer arrivals are made known via an app.

In ride hailing environments, customers do not specify pickup locations but rather hail a nearby passing vehicle.
Ride hailing customer demand estimation has been studied for taxi MOD systems.
In \cite{chang_context-aware_2010,phithakkitnukoon_taxi-aware_2010, li_prediction_2012,davis_multi-level_2016}, temporal demand estimation methods are proposed based on historical data . 
Historical ride data may not be available in autonomous MOD systems if there are no knowledgeable human-taxi drivers to find customers in the first place.
In \cite{shao_estimating_2015, anwar_inferring_2015}, customer demand hotspots are estimated in real-time using large-scale taxi fleets, where the demand is specified either for city-scale areas \cite{shao_estimating_2015} or for customers waiting at taxi stands \cite{anwar_inferring_2015}.
In \cite{ge_energy-efficient_2010,yuan_where_2011,yuan_t-finder:_2012,anwar_inferring_2015}, ride hailing planners are presented in the form taxi recommendation systems.
Taxi-based planners typically focus on recommendations for individual taxi drivers and do not consider fleet-wide coordination available in autonomous systems.

\section{Preliminaries}\label{sec:preliminaries}
\paragraph*{Customer Arrival Model}
This section presents a customer arrival model based on a network graph for an MOD system.
The directed network graph is denoted by $\mathcal{G} = (\mathcal{N},\mathcal{L})$, where $\mathcal{N} = \{ n_1, \dots, n_{N_n} \}$ is a set of $N_n$ nodes.
$\mathcal{L} = \{l_1, \dots, l_{N_l}\}$ is a set of $N_l$ directed link edges each taking the form of an ordered pair of neighbor nodes, $l = (n_{o(l)}, n_{d(l)})\in~\mathcal{N}^2$, where $o(l)$ and $d(l)$ represent the respective origin and destination node indexes of each link.
A route $r(n_{o(r)},n_{d(r)})$ is defined as a sequence of directed links $\mathcal{L}_{r} \subseteq \mathcal{L}$ which corresponds to a unique minimum-travel-time path between origin node $n_{o(r)}$ and destination node $n_{d(r)}$.

Customer demand is quantified as the number of MOD customers that will arrive at nodes within the network graph.
Pedestrians are modeled as arriving in the system with a predetermined route $r$ according to a discrete-time Poisson process with the time-varying arrival rate parameter $\gamma_{r}(t)$.
In this work, the time discretization is sufficiently large such that arrival rates are assumed constant within the considered operating regime, that is $\gamma_{r}(t) = \gamma_r$.
A fraction $f_r$ of total pedestrian route arrivals are potential MOD customers, who will wait briefly at their arrival node to hail a ride.
A waiting customer within proximity of one of the $N_v$ vehicles in the system will receive a ride if the vehicle is unoccupied.
If a customer does not receive a ride after waiting for a time of $t_{wait}$, they will no longer be a potential customer and will walk their route instead.
Due to Poisson superposition and decomposition properties, customer node arrivals occur according to a Poisson process. 
The arrival rate of customers at each node, $\lambda^c_n$, is given by $\lambda^c_n = \sum\limits_{r\in\{r(n,\cdot)\} } f_r \gamma_r$, where $\{r(n,\cdot)\}$ represents the set of routes with $n$ as the origin node.
The number of customer node arrivals, $c_n$, over a time period of $t$, is modeled as $c_n \sim \text{Pois}(c_n; \lambda^c_n \; t)$.

\paragraph*{Bayesian Updates}
This section presents an overview of recursive Bayesian updates for probability distributions utilized in this work.
First, this work measures customer and pedestrian arrivals with uncertain arrival rates.
The number of arrivals, $c$, is modeled as Poisson distributed with uncertain arrival rate parameter $\lambda$ such that $c \sim \text{Pois}(c;\lambda)$
The rate parameter is modeled as Gamma distributed with hyperparameters $\alpha$ and $\beta$ such that $\lambda \sim 
\text{Gamma}(\lambda;\alpha,\beta)$.
Estimation of the arrival rate parameter is made using the number of arrivals, $m$, observed over a time period of $t$. 
The Gamma distribution is conjugate to the Poisson which allows for recursive updates with more observations, that is $\alpha = \alpha_0 + m$ and $\beta = \beta_0 + t$, where $\alpha_0$ and $\beta_0$ are prior values.

Second, this work distinguishes between customer and pedestrian arrivals using binary classification.
A binary quantity $1_c$ is modeled as Bernoulli distributed where $1_c$ indicates a customer with uncertain fraction $p$ and a pedestrian otherwise such that $1_c \sim \text{Bernoulli}(1_c;p)$.
The fraction parameter is modeled as Beta distributed with hyperparameters $a$ and $b$ such that $p \sim 
\text{Beta}(p;a,b)$.
Estimation of the fraction parameter is made using the number of observed ``successes", $c$, and the number of observed ``failures", $\neg c$.
The Beta distribution is conjugate to the Bernoulli which allows for recursive updates with more observations, that is $a = a_0 + c$ and $b = b_0 + \neg c$, where $a_0$ and $b_0$ are prior values.

\section{Customer Arrival Estimation}\label{sec:arrival_estimation}
A Bayesian framework is utilized to estimate customer arrival rates with a quantification of uncertainty.
The key challenge for estimating customer arrivals is the sparsity of observable data.
Direct estimation of the customer arrival rates would require observing the number of customer arrivals at each node over a give period of time.
For ride hailing, this would require vehicles to be placed at each node in order to fully observe the network.
Instead, customer arrival rates are estimated through a two-parameter model which splits customer arrival rates into pedestrian arrival rates and customer fractions in order to utilize more readily observable data.
That is, $\hat{\lambda}^c_n = p_n \lambda_n$, where $\hat{\lambda}^c_n$ is the estimated customer arrival rate, $\lambda_n$ is the pedestrian arrival rate for a node, and $p_n$ is the fraction of pedestrian arrivals which are newly arriving customers.
The benefit to this decomposition is that either large customer fractions or large pedestrian arrival rates can be used as indicators of nodes with large customer arrival rates.
While customer fraction estimation still require that vehicles make observations at the nodes, the pedestrian arrival rates can be estimated from moving vehicles and serve as indicators as to which nodes are worth waiting at.
This allows for identifying spikes in demand by observing unscheduled events that generate high traffic counts. 

Pedestrian arrival rates at nodes are first measured along links in the network graph. 
While pedestrian arrival rates are originally modeled for routes, pedestrians walking routes will also generate Poisson arrivals on links along that route, where link arrival rate parameters, $\mu_l$, are given by $\mu_l = \sum\limits_{r: l \in \mathcal{L}_{r}} \gamma_r$. 
Previous work in \cite{miller_dynamic_2016} presents a moving observer method for estimating link arrival rates using the sensors onboard the MOD vehicles themselves.
The method provides data in the form of a pedestrian link counts $m_l$ with corresponding observation time window $t_l$ for a particular link.
The link arrival rate parameters are modeled and updated using Gamma distributions with hyperparameters, $\hat{\alpha}_l$ and $\hat{\beta}_l$.
Similarly, pedestrians traveling along links will generate arrivals at the origin node of that link, with node arrival rates, $\lambda_n$, given by $\lambda_n = \sum\limits_{l\in\{l(n,\cdot)\} } \mu_l$, where $\{l(n,\cdot)\}$ represents the set of links with $n$ as the origin node.
The node arrival rate parameters are distributed according to a sum of Gamma distributed link arrival rates, and are also modeled using Gamma distributions with hyperparameters, $\alpha_n$ and $\beta_n$.
Node hyperparameters are computed from the link hyperparameters using the the Welch-Satterthwaite approximation for the sum of Gamma distributions provided in \cite{massey_approximation_????}.

Due to the Poisson superposition and decomposition properties, pedestrians node arrivals will be proportioned between new customer route arrivals and arrivals for pedestrians passing through the node.
The customer probabilities at each node are represented using a Bernoulli distribution with uncertain customer fraction parameter $p_n$.
Observations include the number of customers that were picked up at a node, $c_n$, and the number of non-customer pedestrians observed at the node using vehicle sensors, $\neg c_n$.
Customer fractions are modeled and updated using Beta distributions with hyperparameters $a_n$ and $b_n$.

The probability of the number of customer arrivals over a time period of $t_{pred}$ is determined through marginalization of the pedestrian arrival rate and customer fraction parameters.
The marginalization can be simplified into an analytical expression, the derivation of which is provided in \Cref{sec::marginalization}.
The analytical expression is given as
\begin{align}
&P(\hat{c}_n;t_{pred},\alpha_n,\beta_n,a_n,b_n,) = \nonumber\\
&\quad\frac{\Gamma(\alpha_n+\hat{c}_n)}{\hat{c}_n!\Gamma(\alpha_n)}
\cdot\frac{\Gamma(a_n+\hat{c}_n)}{\Gamma(a_n)}
\cdot\frac{\Gamma(a_n+b_n)}{\Gamma(a_n+b_n+\hat{c}_n)}
\cdot\left(\frac{t_{pred}}{\beta_n}\right)^{\hat{c}_n}  \nonumber\\
&\quad\cdot \; _2F_1\left(a_n+\hat{c}_n,\; \alpha_n+\hat{c}_n,\; a_n+b_n+\hat{c}_n,\; -\frac{t_{pred}}{\beta_n}\right),
\label{eqn::customer_probability}
\end{align}
where $\Gamma(\cdot)$ represents the Gamma function and $_2F_1(\cdot,\cdot,\cdot,\cdot)$ represents the hypergeometric function.
The customer arrival rate probability in \Cref{eqn::customer_probability} incorporates both the inherent uncertainty in the Poisson arrival process as well as the parameter uncertainty from online estimation of pedestrian arrival rates and customer fractions.

%The predicted probability of the number of customer arrivals over a time period of $t_{pred}$ is determined through marginalization of the pedestrian arrival rate and customer fraction parameters, that is,
%\begin{align}
%	&P(\hat{c}_n;t_{pred},\alpha_n,\beta_n,a_n,b_n,) = \nonumber\\
%	&\quad\int_{0}^{1}\int_{0}^{\infty}
%	\text{Pois}(p_n \lambda_n t_{pred}) \cdot
%	\text{Gamma}(\alpha_n,\beta_n) \nonumber\\ 
%	&~~\quad\quad\quad\quad \cdot \text{Beta}(a_n,b_n) \; d{\lambda_n} d{p_n}.
%\end{align}
%The marginalization is not trivial but can be simplified to an analytical expression, the derivation of which is omitted for brevity. 
%\jmXX{add appendix}
%The analytical expression is given as
%\begin{align}
%	&P(\hat{c}_n;t_{pred},\alpha_n,\beta_n,a_n,b_n,) = \nonumber\\
%	&\quad\frac{\Gamma(\alpha_n+\hat{c}_n)}{\hat{c}_n!\Gamma(\alpha_n)}
%	\cdot\frac{\Gamma(a_n+\hat{c}_n)}{\Gamma(a_n)}
%	\cdot\frac{\Gamma(a_n+b_n)}{\Gamma(a_n+b_n+\hat{c}_n)}
%	\cdot\left(\frac{t_{pred}}{\beta_n}\right)^{\hat{c}_n}  \nonumber\\
%	&\quad\cdot \; _2F_1\left(a_n+\hat{c}_n,\; \alpha_n+\hat{c}_n,\; a_n+b_n+\hat{c}_n,\; -\frac{t_{pred}}{\beta_n}\right),
%	\label{eqn::customer_probability}
%\end{align}
%where $\Gamma$ represents the Gamma function and $_2F_1$ represents the generalized hypergeometric function.
%The customer arrival rate probability in \Cref{eqn::customer_probability} incorporates both the inherent uncertainty in the Poisson arrival process as well as the parameter uncertainty from online estimation of pedestrian arrival rates and customer fractions.

\section{Exploration to Reduce Uncertainty}\label{sec:exploration}
An exploration planner is proposed for routing vehicles in order to reduce uncertainty in pedestrian arrival rates.
Both pedestrian arrival rates and customer fractions are used to estimate customer arrival rates.
Customer fraction estimation requires vehicles to be near nodes, which can lead to missed customers if vehicles are simply waiting at nodes with low customer arrival rates.
On the other hand, pedestrian arrival rate estimation allows vehicles to continue to explore the network graph as estimation is performed as vehicles traverse links.
For the same customer fraction, nodes with higher pedestrian arrival rates will result in higher customer arrival probability. 
This exploration planner focuses on routing vehicles along links that will most reduce the variance in pedestrian arrival rates.

\subsection*{Problem Formulation}
The belief of the link arrival rate, $\mu_l$, is expressed through the Gamma distribution hyperparameters $\hat{\alpha}_l$ and $\hat{\beta}_l$.
The mean and variance for a Gamma distribution are known to be $\mathbb{E}\left[\mu_l\right] = \hat{\alpha}_l/\hat{\beta}_l$ and $\text{Var}\left[\mu_l\right] = \hat{\alpha}_l/\hat{\beta}^2_l$, respectively.
The expected amount of time to traverse a link is given by $\hat{t}_l = d_l/s_l$, where $d_l$ is the length of the link and $s_l$ is the expected vehicle speed along the link.
The number of pedestrians that are expected to be observed during that time is $\hat{m_l} = \mathbb{E}\left[\mu_l\right] \hat{t_l}$.
Using the Bayesian update for link arrival rates, the expected variance in the link after a traversal is given by
\begin{equation}
	\text{Var}(\hat{\mu_l}) 
	= \frac{\hat{\alpha}_l + \hat{m_l}}
	{(\hat{\beta}_l + \hat{t_l})^2}
	= \frac{\hat{\alpha}_l}
	{\hat{\beta}_l (\hat{\beta}_l + \hat{t}_l)}.
\end{equation}

The exploration problem is that of assigning vehicles to links such that the reduction in link arrival rate variance is maximized.
Vehicles are assigned a route composed of links, where $\Tau$ is the total number of assigned links and $\tau$ represents the link index within the route.
The problem can be formed as a non-linear integer problem with binary decision variables $x^v_{l\tau} \in \{0,1\}$ equal to 1 if vehicle $v$ is assigned to link $l$ for the index $\tau$ within the route; and zero otherwise.
The exploration problem formulation is,
\begin{flalign}
\argmax_{x^v_{l\tau}}  &\quad\sum_{l=1}^{N_l}
\quad\frac{\hat{\alpha}_l}{\hat{\beta}^2_l} -
\frac{\hat{\alpha}_l} {\hat{\beta}_l \left(\hat{\beta}_l + \hat{t}_l \sum\limits_{v=1}^V \sum\limits_{\tau=1}^\Tau x^v_{l\tau}\right)}
\label{eqn::objective}\\
s.t.  &\quad \sum_{\tau=1}^\Tau x^v_{l\tau} = 1 
\quad\quad\quad\quad\quad\quad\quad\quad\quad\forall \; v,l  
\label{eqn::vehicles_assigned}\\
&\quad d(l) \; x^v_{l\tau-1} = o(l) \; x^v_{l\tau} 
\quad\quad\quad\quad\quad\forall \; v, l, \tau,
\label{eqn::continutiy}
\end{flalign}
where $o(l)$ and $d(l)$ represent the known origin and destination node indexes of each link, respectively; and $x^v_{l0}$ for $\tau=0$ is known using each vehicle's current link.
The objective in \Cref{eqn::objective} computes the total number of visits for each link and measures the expected reduction in uncertainty across all links.
\Cref{eqn::vehicles_assigned} ensures that each vehicle is assigned a link for each index in its route.
\Cref{eqn::continutiy} ensures that route continuity is maintained by ensuring that the origin node of the link at $\tau$ is the same as the destination node of the previous link at $\tau-1$.

\subsection*{Online Approach}
\begin{figure}
	\center
	\includegraphics[width=1\columnwidth] {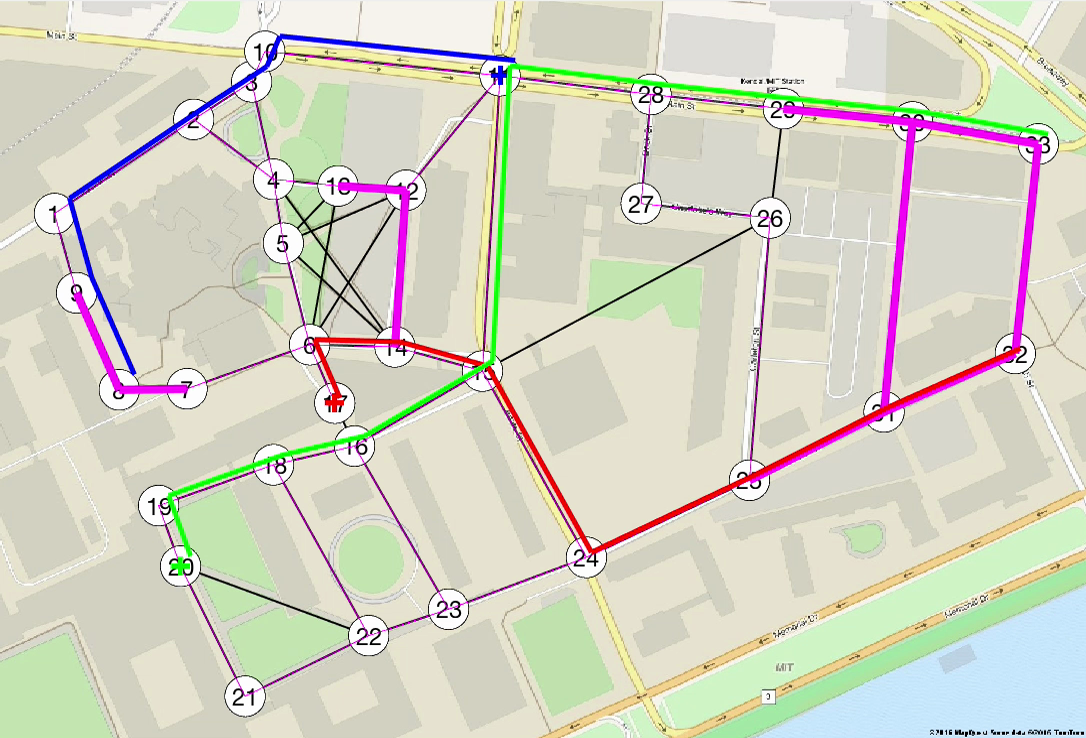}
	\caption{Route assignments from the exploration planner. Relative link uncertainties are shown in maroon. Three vehicles (red, blue, green) are each assigned a route to reduce uncertainty. The route origin for each vehicle is marked with a cross.The traffic network for MIT campus is composed of 33 nodes, 106 directed links, and 1056 precomputed routes.}
	\label{fig::route_assignments}
\end{figure}

The exploration problem can be solved to assign vehicle routes; however, the objective function in \Cref{eqn::objective} is nonlinear and challenging to solve.
To address this, route assignments are determined sequentially for each individual vehicle.
The feasibility constraints in \Cref{eqn::continutiy} are ensured by assigning routes chosen from the set of routes that are pre-computed for each node in the network graph.
%For vehicle $v$, the route decision variable $x^v_{l\tau}$ is determined for each route with length $\tau$ and origin at the vehicle's current node.
Each route is then evaluated under the nonlinear objective in \Cref{eqn::objective}.
The process is repeated for each unassigned vehicle with knowledge of the previous vehicle's route so that future link visits are accounted.
\Cref{fig::route_assignments} shows an example set of route assignments.
A greedy assignment is used so that vehicles are continually assigned new routes whenever their previous route is completed.
Otherwise, routes would have to be recomputed for all vehicles whenever one vehicle completes its route, or vehicles would have to wait until all routes are completed.

\section{Ride Hailing Fleet Management}\label{sec:ride_hailing}
Ride hailing planning frameworks are proposed to assign the number of vehicles at each node to match the number of customer arrivals.
The number of customer arrivals at each node is uncertain, so the ride hailing planners utilize the probability distribution for the customer arrivals presented in \Cref{sec:arrival_estimation}.

\subsection*{Problem Formulation}
Let $c_n$ be the number of customers arriving at node $n$ in a time period of $t_{pred}$.
Let $v_n$ be the number of vehicles assigned to node $n$, where $N_v$ is the total number of available vehicles.
For a given vehicle assignment, the cost, $C_n$, at each node is $C_n = (c_n - v_n)^2$.
%In the case where there are fewer customer arrivals than vehicles in the system, excess vehicles at a node are penalized in order to have more vehicles available for exploration.
%In the case where there are more customer arrivals than vehicles, vehicles are assigned proportionally to the number of customer arrivals.
A quadratic cost function is chosen so that 1) in the case where there are fewer customer arrivals than vehicles in the system, excess vehicles at a node are penalized in order to have more vehicles available for exploration; and 2) in the case where there are more customer arrivals than vehicles, vehicles are assigned proportionally to the number of customer arrivals.
The problem can be formulated as an integer quadratic program with $v_n \in \{1, \dots, N_v\}$ as the decision variables.
The ride hailing problem formulation is,
\begin{flalign}
\argmin_{v_n}  &\quad\sum_{n=1}^{N_n} (c_n - v_n)^2 \label{eqn::rh_objective}\\
s.t.  &\quad\sum_{n=1}^{N_n} v_n \le N_v. \label{eqn::vehicle_constraint}
\end{flalign}
\Cref{eqn::vehicle_constraint} ensures that vehicle assignments do not exceed the fleet size.
The challenge for the problem formulation stems from the uncertainty in the number of customer arrivals at each node.
The formulation is adapted to address the uncertainty by using expected value and chance-constrained approaches.

The expected value formulation assigns vehicles based on the expected number of customers.
The expectation is taken over all customer arrivals changing the objective function in \Cref{eqn::rh_objective} to be
\begin{flalign}
\argmin_{v_n} &\quad\mathbb{E}\left[\sum_{n=1}^{N_n} (c_n - v_n)^2\right]. \label{eqn::ev_objective}
\end{flalign}
Because the probability of customer arrivals at each node is independent, \Cref{eqn::ev_objective} can be rewritten as
\begin{equation}
	\argmin\limits_{v_n} \sum\limits_{n=1}^{N_n} (\mathbb{E}[c_n] - v_n)^2 , \label{eqn::ev_objective_revised}
\end{equation}
where $\mathbb{E}[c_n]$ is the expectation according to \cref{eqn::customer_probability}.

The chance-constrained formulation bounds the total vehicle assignment cost to be within a threshold.
The chance constraint is applied to the objective, changing \Cref{eqn::rh_objective} to be
\begin{flalign}
\argmin_{v_n}  &\quad X \label{eqn::cc1_objective}\\
s.t.  
&\quad \text{P}\left(\sum_{n=1}^{N_n} (c_n - v_n)^2 \le X\right) \ge \eta \label{eqn::cc1_constraint}
\end{flalign}
where $X$ is a decision variable representing the total cost and $\eta$ is a predetermined risk tolerance threshold.
The additional constraint in \Cref{eqn::cc1_constraint} uses the joint probability of the sum of random customer arrivals, which is difficult to compute.
Instead, a more constrained version of the problem is formulated, where the bound is made for each node cost.
The node chance-constrained formulation replaces the objective in \Cref{eqn::rh_objective} with
\begin{flalign}
\argmin_{v_n}  &\quad \sum_{n=1}^{N_n} X_n \label{eqn::cc2_objective}\\
s.t.  
&\quad \text{P}\left( (c_n - v_n)^2 \le X_n \right) \ge \eta_n \label{eqn::cc2_constraint}
\end{flalign}
where $X_n$ are decision variables representing the cost incurred at each node and $\eta_n$ are risk tolerance thresholds for each node.
\Cref{eqn::cc2_constraint} constrains the cost at each node according to the customer arrival probability and can be can be rewritten as,
\begin{align}
	\nonumber\\
	&\quad\quad\text{P}\left( (c_n - v_n)^2 \le X_n \right) \ge \eta_n \nonumber\\
	&\text{P}\left(v_n - \sqrt{X_n} \le c_n \le v_n + \sqrt{X_n} \right) \ge \eta_n \nonumber\\
	&\text{F}\left(v_n + \sqrt{X_n}\right) - \text{F}\left(v_n - \sqrt{X_n}\right) \ge \eta_n, \label{eqn::cc2_constraint_cdf}
\end{align}
where F$(\cdot)$ represents the cumulative distribution function (CDF) for the customer arrivals, computed from \cref{eqn::customer_probability}.

\subsection*{Online Approach}
The ride hailing planner operates on a fixed planning horizon of length $t_{pred}$.
At the start of the planning horizon, the number of vehicles to assign to each node is determined.
After determining the number of vehicles to assign, actual vehicle assignments are made using greedy assignment.
Vehicles are continually assigned over the course of the time horizon to satisfy the determined node assignment numbers.
For example, if a vehicle encounters a customer and leaves an assigned node to serve them, any unassigned vehicles in the system will be assigned to take its place.
The main challenge for the online approach is to determine the number of vehicles to assign to each node.

For the expected value ride hailing planner, the expected number of customers arrivals at each node for $t_{pred}$ is first computed.
Rather than compute the expectation from \cref{eqn::customer_probability}, an iterated expectation is computed over the parameters.
The expectation is given as,
\begin{align}
	\mathbb{E}[c_n] 
	&= \mathbb{E}[p_n] \mathbb{E}[\lambda_n] = \frac{a_n}{a_n+b_n} \frac{\alpha_n}{\beta_n},
\end{align}
where $\mathbb{E}[p_n]$ and $\mathbb{E}[\lambda_n]$ are known for Beta and Gamma distributions, respectively.
\Cref{eqn::ev_objective_revised,eqn::vehicle_constraint} are then solved using integer quadratic programming. 

For the chance-constrained ride hailing planner, the problem is not as easily solved because \Cref{eqn::cc2_constraint_cdf} introduces a nonlinear constraint.
Instead, the problem is decomposed into two sub-problems.
The first problem determines the minimum cost for each vehicle at each node, represented by the cost matrix $K \in R^{N_v \times N_n}$.
The problem is formulated as,
\begin{flalign}
&K(v_n,n) = \argmin_{X_n}  \quad X_n \label{eqn::cc3_objective}\\
&s.t.  
\quad \text{F}\left(v_n + \sqrt{X_n}\right) - \text{F}\left(v_n - \sqrt{X_n}\right) \ge \eta_n. \label{eqn::cc3_constraint} 
\end{flalign}
$K$ is computed through enumeration over all nodes and number of vehicles.
The second problem determines the optimal number of vehicles to assign to each node to minimize the total cost.
The problem is formulated as,
\begin{flalign}
\argmin_{v_n} &\quad\sum_{n=1}^{N}K(v_n,n) \label{eqn::cc4_objective}\\
s.t.  
&\quad\sum_{n=1}^{N_n} v_n \le N_v, \label{eqn::cc4_vehicle_constraint}
\end{flalign}
which is solved using integer linear programming.
One challenge for the chance-constrained planner is that of determining the appropriate risk allocation thresholds for each node.
External knowledge of the MOD network could be used to allocate more risk to certain nodes than others.
In this work, a uniform risk allocation is used where each node is assigned the same risk tolerance value.

\subsection*{Planner Comparison}
\begin{figure}[t]
	\centering
	\begin{subfigure}{0.5\textwidth}
		\centering
		\includegraphics[width=1\columnwidth]{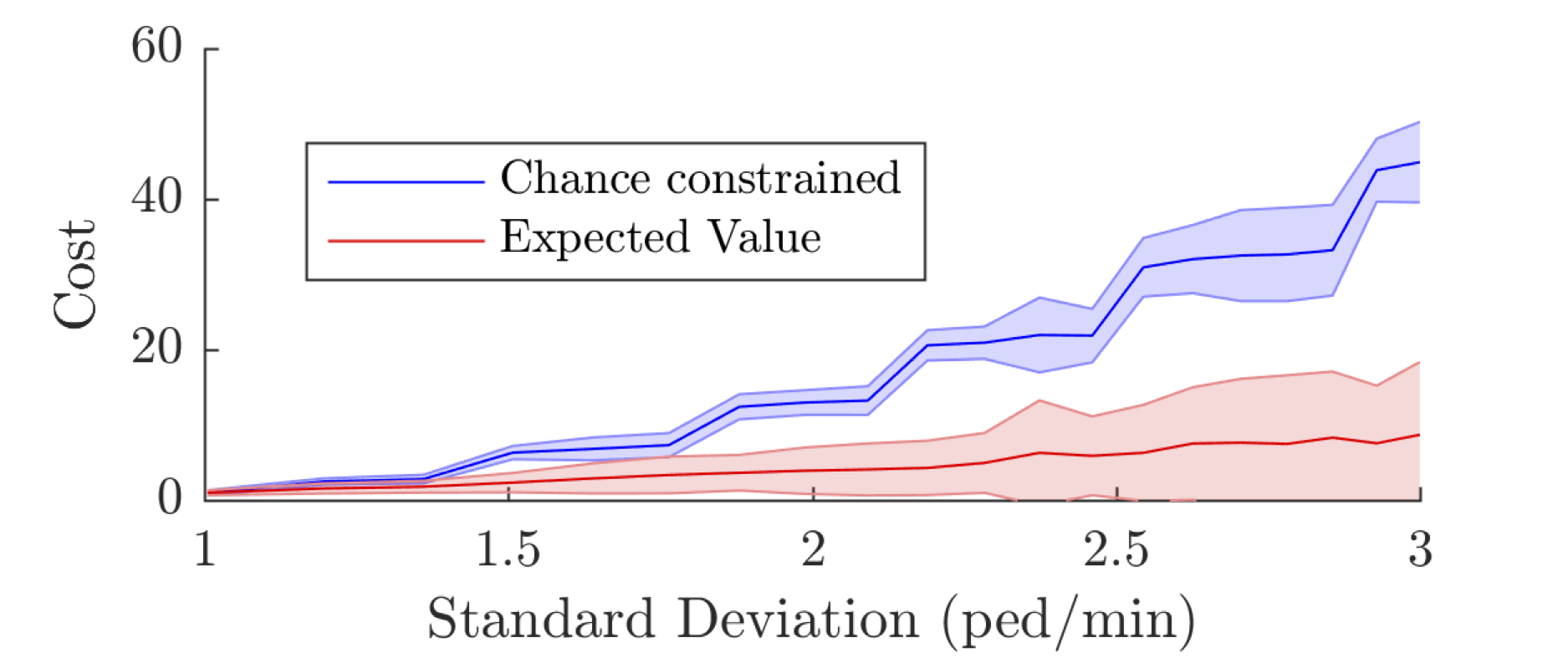}
		\caption{Mean cost}
		\label{fig::cc_vs_ev_mean}
	\end{subfigure}\\
	\begin{subfigure}{0.5\textwidth}
		\centering
		\includegraphics[width=1\columnwidth]{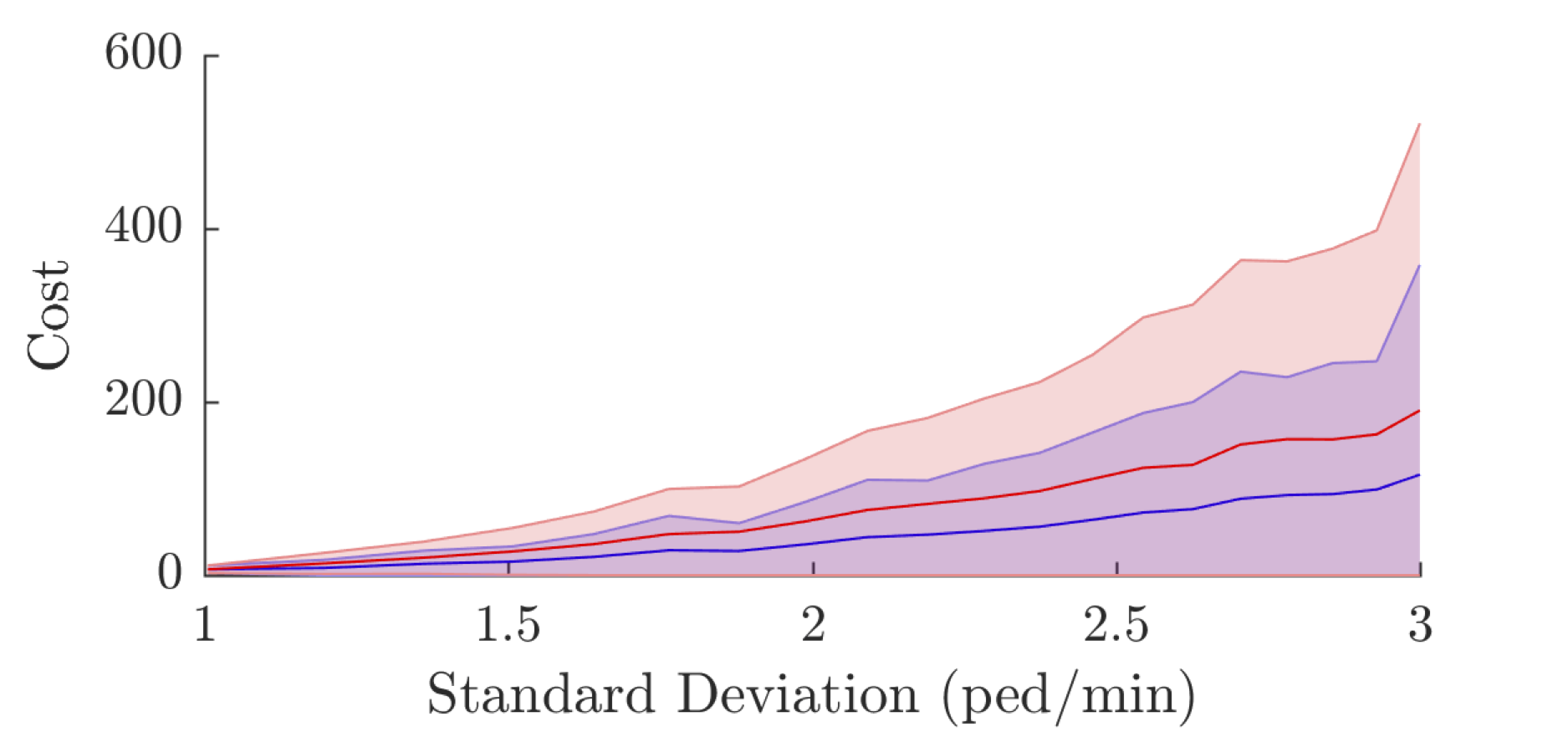}
		\caption{Maximum cost}
		\label{fig::cc_vs_ev_max}
	\end{subfigure}
	\caption{Comparison of expected value and chance-constrained planners. Arrivals are generated from a negative binomial distribution, with mean of 1 ped/min at each node and varying standard deviation. Cost is measured as (a) mean cost or (b) maximum cost across all nodes, lower is better. Note that maximum costs are an order of magnitude higher than the mean costs. Mean and standard deviation are shown for 1000 simulated arrivals.}
	\label{fig::cc_vs_ev}
\end{figure}

The expected value and chance-constrained formulations both utilize the belief in the number of customer arrivals. 
The expected value formulation assigns vehicles based on the expectation of the belief, and is robust only in the sense that the planner should perform well on average.
The chance-constrained formulation uses the full posterior information of the belief through the CDF and allows the cost at each node to vary according to the risk threshold for that node.
To develop a general understanding of the difference between the two planners, a simulation that allows for any number of vehicles to be assigned is performed, where the chance-constrained risk tolerance is set to $\eta_n=0.99$ to better emphasize the difference.
Poisson customer arrivals for a 33 node network graph are generated from a negative binomial approximation, where each node has mean arrival rate of 1 ped/min but with a set variance.
Two performance metrics are studied.
First, is the mean cost over all nodes, that is $\bar{C} = \frac{1}{N_n} \sum\limits_{n=1}^{N_n} C_n$.
Second, is the maximum cost for any single node, that is $C_{max} = \max (C_1, \dots, C_{N_n})$.
\Cref{fig::cc_vs_ev} shows comparisons of expected value and chance-constrained planners under increasing amounts of variance.
The expected value planner has lower mean cost because it assigns the correct number of vehicles on average, with no consideration towards potentially high node costs.
The chance-constrained planner has lower maximum cost because it assigns more vehicles in order to bound any potentially high individual node cost, which comes at the price of higher mean cost.
The choice of planner ultimately comes down to the goal of the MOD system, whether it is more important to perform well on average or to bound performance on any individual node.

\section{Experiments}\label{sec:experiments}
The ride hailing fleet management framework is evaluated using data-driven simulation of the real-world MIT MOD system.
There are two motivating test cases: 1) evaluating how well the planners can learn customer arrival rates, and 2) evaluating how well the planners can increase the number of served hailing customers.
The MIT MOD system is used to provide simulation parameters that reflect a realistic operating environment for vehicles and customers.

\subsection*{Simulation Setup}
Pedestrians and vehicles operate within a network graph for the MIT campus.
The network graph, shown in \cref{fig::route_assignments}, is generated using pedestrian trajectory data collected from sensors onboard the MOD vehicles, following the work presented in \cite{miller_dynamic_2016}.
Pedestrian arrivals are sampled from a Poisson process with route arrival rates determined to reflect values presented in \cite{miller_dynamic_2016}.
Not all routes will have pedestrian or customer arrivals.
Ten randomly chosen routes with separate origin nodes are assigned pedestrian arrival rates of 1 ped/min and three of the ten routes will have all customer arrivals.
Customers wait at their origin node and will successfully hail a ride if a vehicle comes within a 20 meter proximity.
If no vehicle arrives within 30 seconds, the customer is not served and will instead walk their route.
In this simulation, customers can only be picked up at their origin nodes in the network graph and can not be served after they decide to walk.
The simulation considers 5 vehicles in the MIT MOD system.
Both vehicles and walking pedestrians travel along links in the network graph according to their respective velocities. 
Pedestrians link speeds are nominally 1.5 m/s and vehicle link speeds are either 11 m/s or 4 m/s for links corresponding to either city streets or shared pedestrian pathways, respectively.
A period of 1 hour is simulated.

Four different ride hailing planners are analyzed.
Each ride hailing planner operates on a 5 minute planning horizon.
At the start of the planning horizon, the number of vehicles to assign to each node is first determined based on the ride hailing planner.
If fewer vehicles are needed than there are vehicles in the system, the remaining vehicles are assigned to drive routes according to the exploration planner, where assigned routes have a length of 5 links.
In any of the planners, if a vehicle encounters a hailing customer, the vehicle immediately serves the customer.
The first two planners are the expected value and chance-constrained planners formulated in \Cref{sec:ride_hailing}. 
The third is a sensing planner which represents a baseline approach where vehicles continually patrol the network in order to happen upon a hailing customer.
The planner is implemented by never assigning vehicles to wait at nodes so that they are always assigned to traverse links through the exploration planner.
Finally, the fourth is an oracle planner designed to represent the upper bound on performance for serving customers.
The planner is implemented by providing the true number of customer arrivals over the planning horizon to the expected value planner, causing vehicles to often be assigned to wait at nodes.
The oracle is provided only with the actual number of arrivals, and  not the true arrival rate parameters, so that estimation performance can still be tested. 

\begin{figure}
	\center
	\includegraphics[width=1\columnwidth] {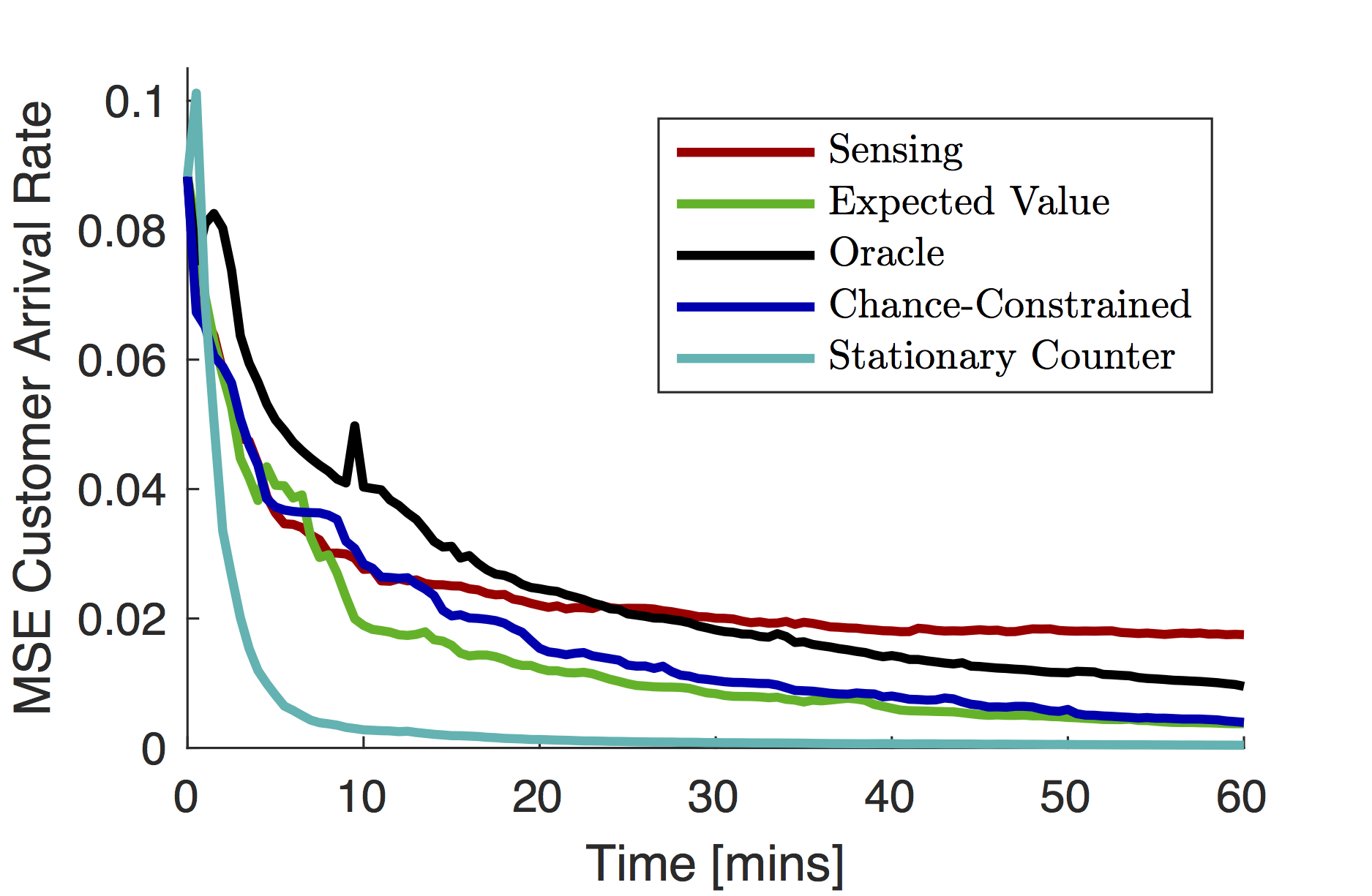}
	\caption{Customer arrival rate estimation error as a function of time. The error is computed as the mean-squared error between the true and estimated customer arrival rates, where less error is better. Each line is the mean error profile over 100 runs.}
	\label{fig::rate_convergence}
\end{figure}

\subsection*{Exploration}
In campus MOD systems, arrival rates can vary throughout the day, so it is important to determine the timescales for which arrival rates can be estimated using the exploration and ride hailing planners.
The simulation considers all planners initially having a uniform uninformative prior belief for each node, and evaluates the mean squared error between the true and estimated customer arrival rates over the course of 1 hour.
In addition to the four strategies above, a stationary counter approach is considered to provide the best-case scenario in rate estimation performance.
The stationary counter approach is implemented by monitoring the true number of pedestrian and customer arrivals at every node in the network graph, the practical implementation of which would be costly.
\Cref{fig::rate_convergence} shows the performance of each strategy for the one hour operation.
Initially, the prior beliefs result in high starting error.
The sensing, chance-constrained, and expected value planners initially assign all vehicles to explore the network.
During the first 10 minutes, the arrival rate error is reduced from exploration.
After that time, customer fraction estimates are improved by the chance-constrained and expected value planners which have begun to assign vehicles to nodes.
The sensing planner rarely improves customer fraction estimates as vehicles are never assigned to wait at nodes, thus the error converges to a higher value.
The oracle planner immediately assigns vehicles to known arrivals and is slower to reduce error as vehicles only explore the network graph to serve customers.
The results demonstrate the effectiveness of the exploration and ride hailing planners for improving customer arrival rate estimates over time.
For estimation of arrival rates which would change on short time scales (10 minutes), the exploration planner performs better than waiting.
For changes over larger time scales (1 hour), the estimation accuracy approaches that of the lower-bound stationary counter.

\begin{figure}[t]
	\centering
	\begin{subfigure}{0.5\textwidth}
		\centering
		\includegraphics[width=1\columnwidth]{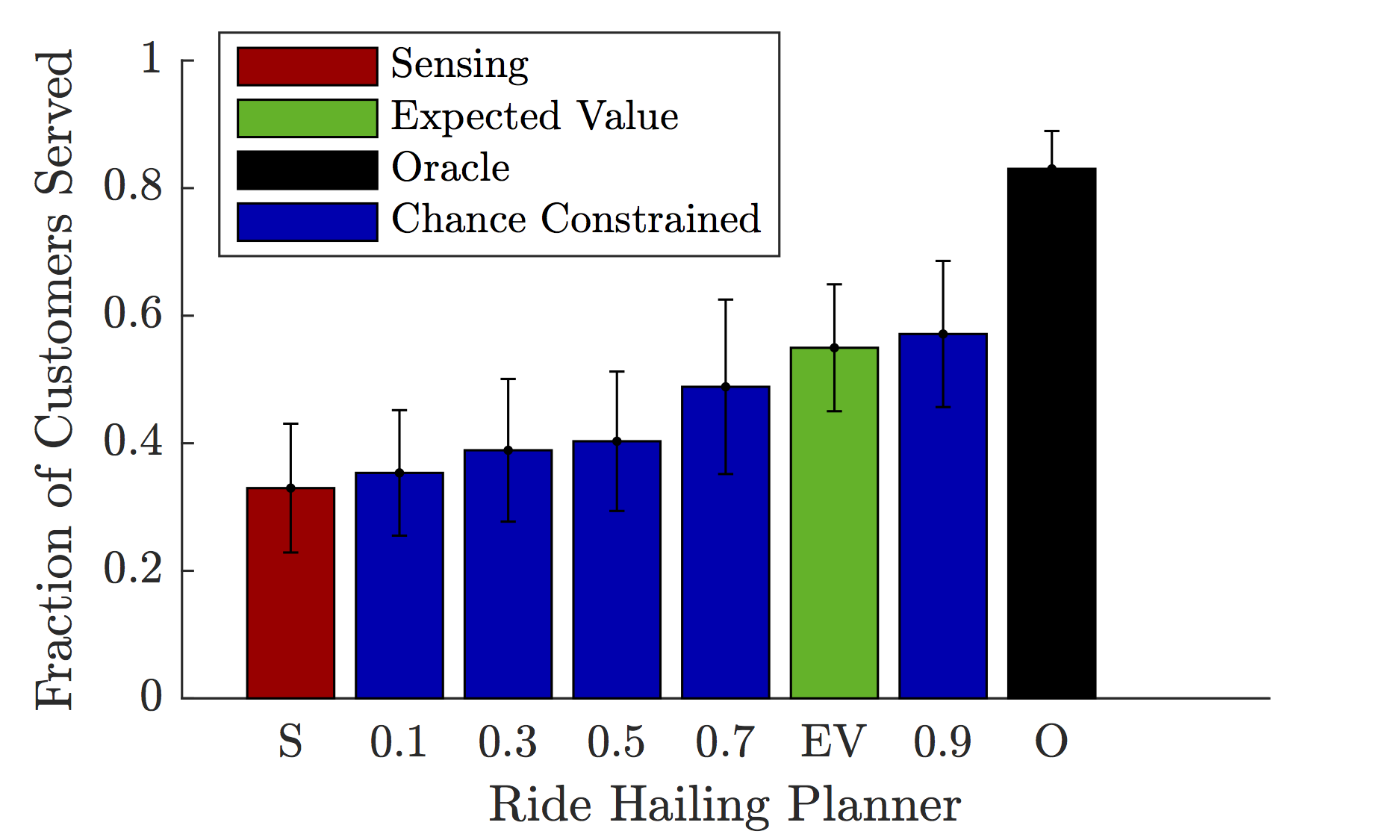}
		\caption{Overall fraction of customers served}
		\label{fig::cc_eta_perf}
	\end{subfigure}\\
	\begin{subfigure}{0.5\textwidth}
		\centering
		\includegraphics[width=1\columnwidth]{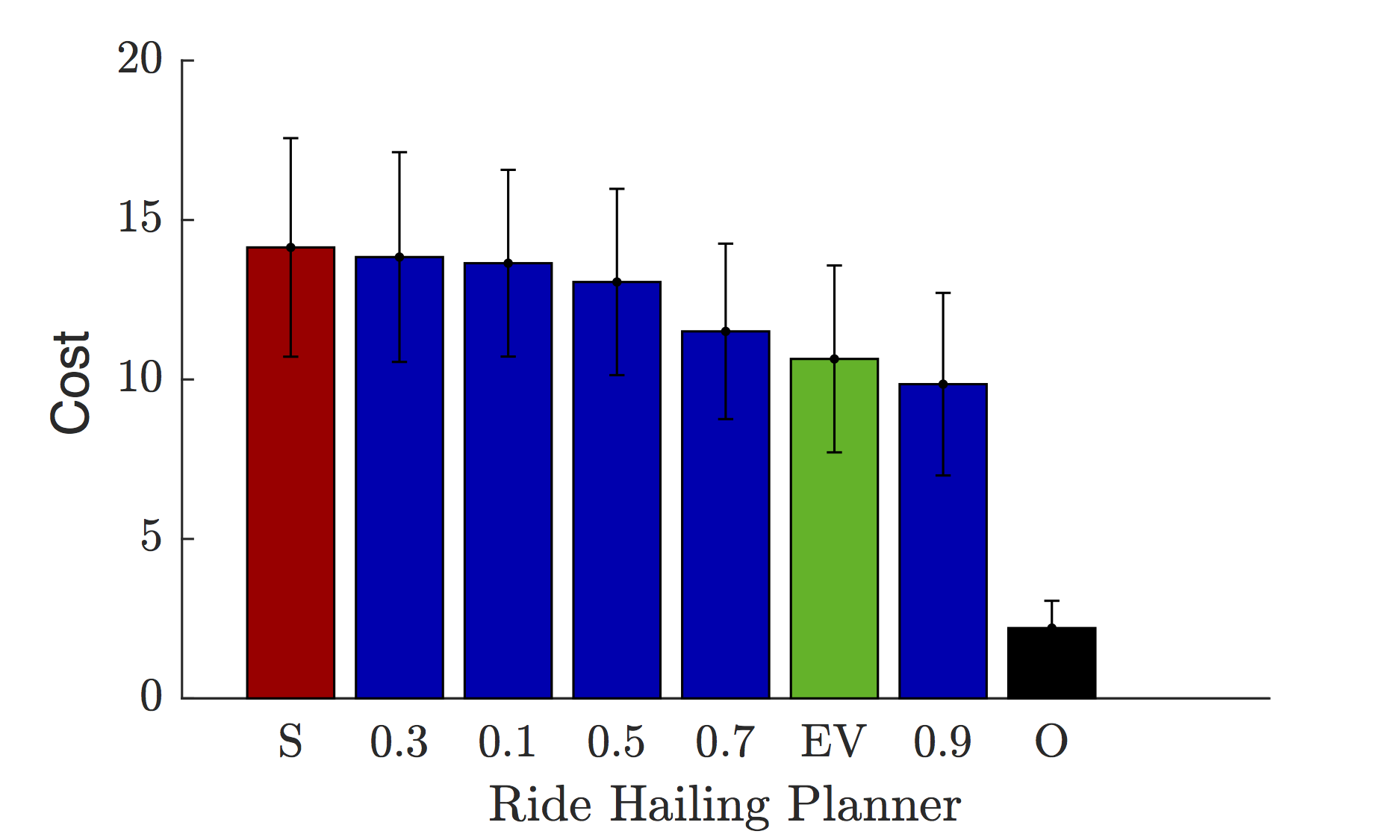}
		\caption{Maximum cost across time horizons}
		\label{fig::cc_eta_cost}
	\end{subfigure}
	\caption{Comparison of sensing, expected value, oracle, and chance-constrained planners. The chance-constrained strategy is compared with several risk allocation values as indicated by the x-axis. (a)~Performance is measured the fraction of customer population that is served, higher is better. (b) Cost is measured as the highest cost of any time horizon, lower is better. Mean and standard deviation shown for 100 simulated runs.}
	\label{fig::cc_eta}
\end{figure}

\subsection*{Ride Hailing}
The performance of the ride hailing planners is evaluated in terms of two performance metrics: the average fraction of customers served for the whole network, and the average highest cost for any individual planning horizon.
With the chance-constrained formulation, uncertainty is managed according to a risk tolerance parameter for each node $\eta_n$.
To determine the effect of the risk tolerance parameter, several chance-constrained planners with varying risk are evaluated along with the other planners. 
\Cref{fig::cc_eta} show the two performance metrics for each of the planners.
First, the results indicate that a higher risk tolerance generally leads to better performance with $\eta_n=0.9$ performing best.
Second, the performance of the oracle planner demonstrates the difficulty of the problem, in that a cost is still incurred due to there being more customer arrivals than vehicles available.
Third, the chance-constrained planner demonstrates an advantage over the expected value planner with a 4\% improvement faction of customers served and 7\% reduction in maximum cost.
Finally, the chance-constrained planner significantly outperforms the baseline sensing planner with a 73\% improvement in the faction of customers served and a 30\% reduction in maximum cost.
This is due to the fact that, in the presence of unknown customer arrival locations, the chance-constrained ride hailing planner successfully learns and utilizes customer demand to improve the number of customers that are served.

\section{Conclusion}\label{sec:conclusion}
 This paper presented an MOD fleet management framework which addresses the challenges of learning and utilizing customer demand for ride hailing environments.
Ride hailing is of particular importance for campus MOD systems, where customers who have the option of walking may be unwilling to request rides through other means such as waiting at stations or using an app.
To address the challenge of customer demand not being specified with ride hailing, a customer demand model is presented which utilizes observable pedestrian arrival rates to estimate customer arrival locations.
Two fleet management planners were presented.
An exploration planner routes vehicles throughout an MOD network in order to reduce uncertainty in pedestrian arrival rates.
A robust chance-constrained ride hailing planner assigns vehicles to wait at customer arrival locations which are subject to arrival rate uncertainty.
Data-driven simulation of the MIT MOD system was used to validate the performance of the planners.
The customer demand model and exploration planner were demonstrated to reduce estimation error over time.
The chance-constrained ride hailing planner was shown to improve the fraction of served customers in the system by 73\% over the baseline exploration approach.

Future work will apply these techniques to the physical MIT MOD system where MIT students will be able to hail rides from the shuttles shown in \Cref{fig::fleet}.
In particular the appropriateness of the Poisson model for customer arrivals will be studied, with emphasis on capturing and predicting short-term bursts in arrivals.
The modeling approach presented in this work can be extended to non-Poisson arrival models; however, new techniques for efficient computation may be needed.
Additionally, the uninformed priors chosen for this work can be improved by extracting temporal arrival patterns from pedestrian and customer data collected over long-term MOD operation.
 
\section*{Acknowledgment}
Research supported by the Ford Motor Company through the Ford-MIT Alliance.
 
 \balance
 
\bibliographystyle{IEEEtran}
\bibliography{IROS2017}

%%%%%%%%%%%%%%%%%%%%%%%%%%%%%%%%%%%%%%%%%%%%%%%%%%%%%%%%%%%%%%%%%%%%%%%%%%%%%%%%

\pagebreak
\nobalance
\onecolumn
\appendix
\subsection{Marginalization of Arrival Probability} \label{sec::marginalization}

\Cref{sec:arrival_estimation} presents a two-parameter model for estimating customer arrivals based on pedestrian arrival rate parameters, $\lambda_n$, that are modeled using Gamma distributions with hyperparameters, $\alpha_n$ and $\beta_n$, and customer fractions, $p_n$, that are modeled using Beta distributions with hyperparameters $a_n$ and $b_n$.
The probability of the predicted number of customer arrivals, $\hat{c}_n$, over a time period of $t_{pred}$ is determined through marginalization of the pedestrian arrival rate and customer fraction parameters.
The derivation of the analytical expression for the probability of $\hat{c}_n$ is provided as,

\begin{align*}
&P(\hat{c}_n;t_{pred},\alpha_n,\beta_n,a_n,b_n) \nonumber\\
&= \quad\int_{0}^{1}\int_{0}^{\infty} 
P(\hat{c}_n;\lambda_n, p_n, t_{pred}) \cdot P(\lambda_n;\alpha_n,\beta_n) 
\cdot  P(p_n;a_n,b_n)
\; d{\lambda_n} d{p_n}.\\
&=\quad\int_{0}^{1}\int_{0}^{\infty}
\text{Pois}(\hat{c}_n; p_n \lambda_n t_{pred}) \cdot \text{Gamma}(\lambda_n;\alpha_n,\beta_n) 
\cdot \text{Beta}(p_n;a_n,b_n) \; d{\lambda_n} d{p_n} \\
&=\quad\int_{0}^{1} \text{Beta}(p_n;a_n,b_n) \int_{0}^{\infty}
\text{Pois}(\hat{c}_n; p_n \lambda_n t_{pred}) \cdot \text{Gamma}(\lambda_n;\alpha_n,\beta_n) \; d{\lambda_n} d{p_n} \\
&=\quad\int_{0}^{1} \text{Beta}(p_n;a_n,b_n) \int_{0}^{\infty} \frac{ e^{-p_n \lambda_n t_{pred}} (p_n \lambda_n t_{pred})^{\hat{c}_n} }{\hat{c}_n!} \cdot \frac{\beta_n^{\alpha_n} \lambda_n^{\alpha_n-1}  e^{-\lambda_n \beta_n}}{\Gamma(\alpha_n)}  \; d{\lambda_n} d{p_n} \\
&=\quad\int_{0}^{1} \text{Beta}(p_n;a_n,b_n) \cdot \frac{\beta_n^{\alpha_n} \; (p_n t_{pred})^{\hat{c}_n} }{\hat{c}_n! \; \Gamma(\alpha_n)} \cdot \int_{0}^{\infty} e^{-\lambda_n (\beta_n + p_n t_{pred})} {\lambda_n}^{(\alpha_n+\hat{c}_n)-1}  \; d{\lambda_n} d{p_n} \\
&=\quad\int_{0}^{1} \text{Beta}(p_n;a_n,b_n)\cdot \frac{\beta_n^{\alpha_n} \; (p_n t_{pred})^{\hat{c}_n} }{\hat{c}_n! \; \Gamma(\alpha_n)} \cdot \frac{\Gamma(\alpha_n+\hat{c}_n)}{(\beta_n + p_n t_{pred})^{\alpha_n+\hat{c}_n}} \; d{p_n} \\
&=\quad\int_{0}^{1} \frac{p_n^{a_n-1} (1-p_n)^{b_n-1}}{\text{B}(a_n,b_n)} \cdot \frac{\beta_n^{\alpha_n} \; (p_n t_{pred})^{\hat{c}_n} }{\hat{c}_n! \; \Gamma(\alpha_n)} \cdot \frac{\Gamma(\alpha_n+\hat{c}_n)}{(\beta_n + p_n t_{pred})^{\alpha_n+\hat{c}_n}} \; d{p_n} \\
&=\quad \frac{\beta_n^{\alpha_n} \; t_{pred}^{\hat{c}_n} \; \Gamma(\alpha_n+\hat{c}_n)}{\text{B}(a_n,b_n) \; \hat{c}_n! \; \Gamma(\alpha_n)} \cdot \quad\int_{0}^{1} \frac{p_n^{\hat{c}_n + a_n - 1} (1-p_n)^{b_n-1}}{(\beta_n + p_n t_{pred})^{\alpha_n+\hat{c}_n}}  \; d{p_n} \\
&=\quad \frac{\beta_n^{\alpha_n} \; t_{pred}^{\hat{c}_n} \; \Gamma(\alpha_n+\hat{c}_n)}{\text{B}(a_n,b_n) \; \hat{c}_n! \; \Gamma(\alpha_n)} \cdot \beta_n^{-\alpha_n - \hat{c}_n} \; \Gamma(b_n) \; \Gamma(a_n +\hat{c}_n) \cdot \;  _2\tilde{F}_1\left(a_n+\hat{c}_n,\; \alpha_n+\hat{c}_n,\; a_n+b_n+\hat{c}_n,\; -\frac{t_{pred}}{\beta_n}\right) \\
&=\quad \frac{\Gamma(a_n + b_n)}{\Gamma(a_n) \; \Gamma(b_n)} \cdot \frac{ t_{pred}^{\hat{c}_n} \; \Gamma(\alpha_n+\hat{c}_n) \; \Gamma(b_n) \; \Gamma(a_n +\hat{c}_n)}{\beta_n^{\hat{c}_n} \hat{c}_n! \; \Gamma(\alpha_n)} \cdot \; \frac{_2F_1\left(a_n+\hat{c}_n,\; \alpha_n+\hat{c}_n,\; a_n+b_n+\hat{c}_n,\; -\frac{t_{pred}}{\beta_n}\right) } {\Gamma(a_n+b_n+\hat{c}_n)} \\
&=\quad\frac{\Gamma(\alpha_n+\hat{c}_n)}{\hat{c}_n!\Gamma(\alpha_n)}
\cdot\frac{\Gamma(a_n+\hat{c}_n)}{\Gamma(a_n)}
\cdot\frac{\Gamma(a_n+b_n)}{\Gamma(a_n+b_n+\hat{c}_n)}
\cdot\left(\frac{t_{pred}}{\beta_n}\right)^{\hat{c}_n} \cdot \; _2F_1\left(a_n+\hat{c}_n,\; \alpha_n+\hat{c}_n,\; a_n+b_n+\hat{c}_n,\; -\frac{t_{pred}}{\beta_n}\right).
\end{align*}
\vspace{0.5cm}

$\Gamma(\cdot)$ represents the Gamma function, $\text{B}(\cdot,\cdot)$ represents the Beta function, $_2{F}_1(\cdot,\cdot,\cdot,\cdot)$ represents the hypergeometric function, and $_2\tilde{F}_1(\cdot,\cdot,\cdot,\cdot)$ represents the regularized hypergeometric function.

The Gamma and Beta functions are related by 
\begin{equation*}
\text{B}(x,y) = \frac{\Gamma(x) \; \Gamma(y)}{\Gamma(x+y)}.
\end{equation*}

The hypergeometric and regularized hypergeometric functions are related by
\begin{equation*}
	_2\tilde{F}_1(w,x,y,z) = \frac{_2F_1(w,x,y,z)}{\Gamma(y)}.
\end{equation*}

%\listofchanges

\addtolength{\textheight}{-0cm}
\end{document}